\RequirePackage{import}
\documentclass[11pt]{article}
\usepackage[preprint]{acl}
\usepackage{times}
\usepackage{latexsym}
\usepackage[T1]{fontenc}
\usepackage[utf8]{inputenc}
\usepackage{microtype}
\usepackage{inconsolata}
\usepackage{graphicx}
\usepackage{booktabs}
\usepackage{colortbl}
\usepackage{multirow}
\usepackage{amsmath}
\usepackage{amssymb}
\usepackage{xspace}
\usepackage{algorithm}
\usepackage{algorithmic}
\usepackage{array}
\usepackage{newfloat}
\usepackage{placeins}
\usepackage[most]{tcolorbox}
\definecolor{SCtableblue}{RGB}{40,103,160}

\newfloat{listing}{tb}{lst}{}
\floatname{listing}{Listing}
\newtcolorbox[use counter=listing]{promptlisting}[2]{
  enhanced,
  breakable,
  label={#1},
  title={Listing~\thetcbcounter: #2},
  title after break={Listing~\thetcbcounter\ (continued)},
  fonttitle=\normalfont,
  coltitle=black,
  colbacktitle=white,
  colback=white,
  colframe=black,
  boxrule=0pt,
  toprule=0.4pt,
  bottomrule=0.4pt,
  leftrule=0pt,
  rightrule=0pt,
  sharp corners,
  left=0pt,
  right=0pt,
  top=2pt,
  bottom=2pt,
  before skip=8pt,
  after skip=8pt,
  before upper={\small\raggedright}
}

\newcommand{\method}{\textsc{SkillCommit}\xspace}

\title{SkillCommit: Evolving Agent Skills through Behaviorally Validated Scope Expansion}
\author{Yu He\textsuperscript{1} \quad Weikai Yang\textsuperscript{2}\thanks{Corresponding author.} \\
\textsuperscript{1}Nanjing University \\
\textsuperscript{2}The Hong Kong University of Science and Technology (Guangzhou) \\
\texttt{yu\_he@smail.nju.edu.cn}, \texttt{weikaiyang@hkust-gz.edu.cn}}

\begin{document}
\maketitle

\begin{abstract}
Large language model (LLM) agents can continually improve without parameter updates by converting historical experience into reusable procedural knowledge.
However, existing methods often consolidate experience based on semantic similarity or LLM judgments, which may merge superficially related but behaviorally incompatible strategies and thereby degrade performance.
To address the issue, we propose \method, an online skill evolution framework that continuously transforms experience into a hierarchical library of reusable skills.
Rather than generalizing from individual experiences immediately, \method first preserves each experience as an instance-specific patch, retaining its locally validated behavior. 
As patches accumulate, \method abstracts those sharing a common behavioral mechanism into higher-level skills.
Specifically, for each incoming skill, embedding-based retrieval first identifies candidate related skills.
Cross-instance replay and an LLM-based mechanism check determine whether these skills transfer across cases and share a common underlying mechanism.
Candidates that pass both checks are abstracted into a higher-level skill and committed only if it preserves the validated behavior of all constituent skills.
Experiments on RuleArena, OpenExempt and KOR-Bench demonstrate that \method consistently improves agent performance across diverse domains.  %
Moreover, the learned skills transfer across model scales and families, enabling cross-model experience transfer.

\end{abstract}

\section{Introduction}
Large language model (LLM) agents can use external tools and interact with their environments with increasing autonomy \citep{yao2023react,schick2023toolformer,yang2023gpt4tools,deng2023mind2web,zhou2024webarena}.
Yet reliable deployment in specialized domains often requires domain knowledge, procedural conventions, and operational constraints that base models do not acquire during pretraining.
Reusable skills provide a modular way to externalize such procedures and guide agents without modifying model parameters \citep{wang2025agentworkflow,ni2026trace2skill,ma2026skillgen,shinn2023reflexion,zhao2024expel,ouyang2026reasoningbank}.
In practice, however, such knowledge is rarely available as a complete instruction document and instead remains scattered across practitioners' experience, execution histories, and task feedback.
Manually eliciting and formalizing such knowledge is costly and may still omit rare but critical boundary cases.
This gap motivates \textit{skill evolution}, in which agents automatically extract, consolidate, and refine reusable skills from accumulated trajectories and feedback \citep{ouyang2026reasoningbank,zhang2026ace,zhang2026ca3mem,chen2026skillcat,suzgun2026dynamic}.

\begin{figure*}[t]
\centering
\includegraphics[width=0.98\textwidth]{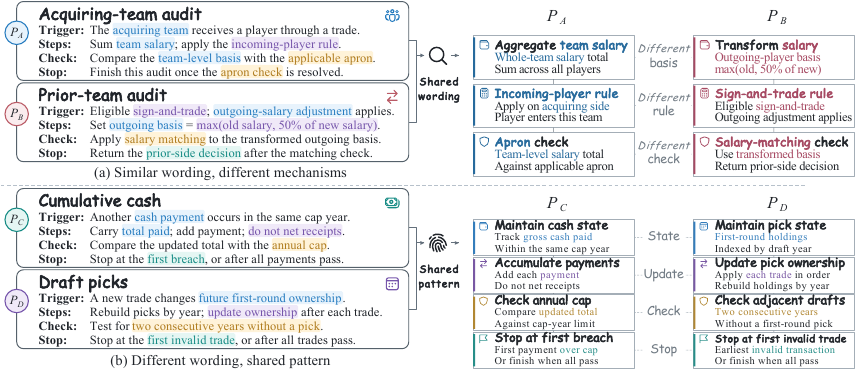}
\caption{
Why semantic similarity alone is insufficient for identifying reusable skills. (a) Similar wording can conceal different behavioral mechanisms. (b) Different wording can describe the same reusable behavioral pattern.
}
\label{fig:semantic-behavior-mismatch}
\end{figure*}

Existing methods typically use LLM-based abstraction to distill individual trajectories and feedback into instance-specific patches that agents can retrieve and reuse when solving similar tasks \citep{shinn2023reflexion,zhao2024expel,ouyang2026reasoningbank}.
However, since these patches are narrow in scope, retaining them individually leads to growing memory and retrieval costs as experience accumulates \citep{chen2026skillcat,ni2026trace2skill,zhang2026ca3mem}.
To build a more compact and general skill library, subsequent methods consolidate related instance-specific patches into higher-level skills based on semantic similarity \citep{xu2025amem,ma2026skillgen} or model judgments \citep{xu2025amem,wang2025agentworkflow,ni2026trace2skill,ma2026skillgen,chen2026skillcat}.
However, they do not guarantee behavioral compatibility, because semantically similar patches may encode conflicting strategies, whereas dissimilar patches may share a common underlying mechanism (\autoref{fig:semantic-behavior-mismatch}).

Reliable skill evolution therefore requires criteria that go beyond surface similarity.
Research in cognitive science suggests that concrete experience must first be transformed into procedural knowledge that can guide future action \citep{kolb2014experiential}.
Such knowledge transfers beyond its original context only when learners identify the relational structure shared across experiences rather than relying on surface resemblance \citep{gentner1983structure,gick1983schema,chi1981categorization}.
Moreover, newly formed abstractions must be integrated without interfering with previously validated competence \citep{mcclelland1995complementary}.
Together, these principles suggest that skill evolution should establish the local validity, justify its broader reuse through evidence of a shared underlying mechanism, and preserve validated behavior as that knowledge is consolidated at higher levels of abstraction.

To meet these requirements, we propose \method, an online skill evolution framework that organizes skill evolution into three stages.
First, when the agent fail to complete a task, \textbf{Feedback-Guided Patch Induction} derives an instance-specific patch based on the failed trajectory and feedback and locally validates it by replaying this task instance.
As patches accumulate, \textbf{Behavioral Compatibility Grouping} evaluates whether multiple patches exhibit transferable behavior and share a common underlying mechanism, thereby determining the appropriate scope of a candidate higher-level skill.
Finally, \textbf{Source-Preserving Consolidation} abstracts behaviorally compatible patches into a higher-level skill and commits it when it preserves the validated behavior of every constituent skill.
This process allows the skill library to expand, reorganize, and compress over time while retaining empirically validated behavior.

We evaluated \method across RuleArena \citep{zhou2025rulearena}, OpenExempt \citep{servantez2026openexempt} and KOR-Bench \citep{ma2025korbench}, where it consistently outperforms no-skill settings and strong baselines across all evaluated subsets.
Further experiments show that the induced skills transfer effectively across model scales and and even across model families.
Our contributions are:
\begin{itemize}
    \setlength{\itemsep}{1pt}
    \setlength{\parsep}{0pt}
    \setlength{\parskip}{0pt}
    \item We formulate reliable skill evolution as behaviorally validated scope expansion across skill admission, compatibility grouping, and higher-level abstraction, requiring replay evidence for every broader behavioral claim.
    \item We propose \method, an online framework to build compact skill libraries while retaining previously validated behavior. %
    \item We evaluate \method across RuleArena, OpenExempt, and KOR-Bench, demonstrating consistent improvements over strong baselines and skill transfer across model scales and families.
\end{itemize}

\section{Related Work}
\paragraph{LLM Agents and Skill Systems.}
Agent skills have emerged as reusable procedural artifacts that extend LLM agents with task instructions, operational knowledge, and executable workflows beyond one-off prompting and atomic tool use \citep{wang2023voyager,wang2025agentworkflow,li2026skillsbench,ni2026trace2skill,ma2026skillgen}.
Prior work has studied how such external knowledge is organized and associated, with A-MEM \citep{xu2025amem} dynamically linking accumulated memory notes and CA3Mem \citep{zhang2026ca3mem} recombining related experience through a structured memory graph.
Agent Workflow Memory \citep{wang2025agentworkflow} instead induces recurring workflows from agent trajectories and selectively retrieves them for subsequent tasks.
Meanwhile, SkillsBench \citep{li2026skillsbench} evaluates whether curated skills improve agent performance, whereas SkillLearnBench \citep{zhong2026skilllearnbench} examines whether agents can continually acquire reusable skills from task experience.
These studies establish how skills can be organized, retrieved, and evaluated, but leave open what behavioral evidence should justify the broader applicability claimed by an evolving skill.

\paragraph{Skill Acquisition and Self-Evolution.}
Early experience-based methods\citep{zhang2024agentpro,gupta2024metareflection} transform execution feedback into reusable textual knowledge, including verbal reflections in Reflexion \citep{shinn2023reflexion}, transferable insights in ExpeL \citep{zhao2024expel}, and reasoning strategies distilled from successful and failed experience in ReasoningBank \citep{ouyang2026reasoningbank}.
More recent systems move beyond retaining individual memories by continually refining an evolving playbook in ACE \citep{zhang2026ace} or hierarchically consolidating trajectory-local lessons in Trace2Skill \citep{ni2026trace2skill}.
SkillGen \citep{ma2026skillgen} evaluates the net intervention effect of a synthesized skill, while SkillCAT \citep{chen2026skillcat} replays candidate patches on their source tasks before hierarchical merging.
Although these methods introduce validation at particular stages, they do not uniformly govern the successive scope expansions from a textual candidate to a validated patch, from individual patches to a behaviorally compatible group, and from a group abstraction to an active pattern.
In contrast, \method validates these decisions through teacher-referenced source replay, directed cross-instance replay, and source-retention replay.

\section{SkillCommit}
\begin{figure*}[t]
\centering
\includegraphics[width=\textwidth]{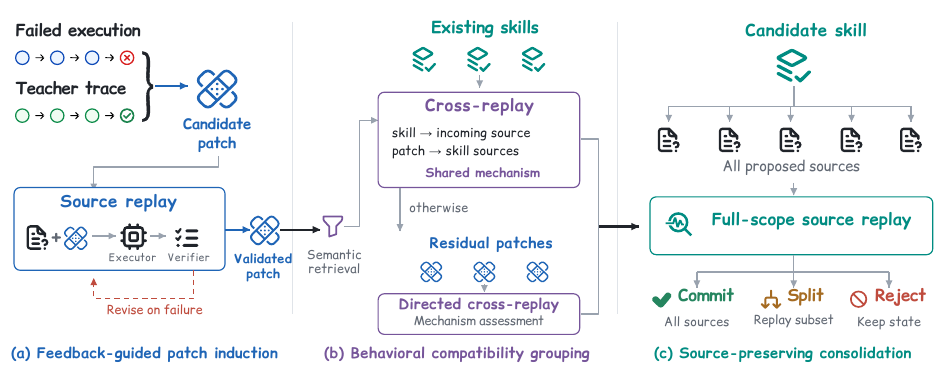}
\caption{\method overview. (a) Induce and validate instance-specific patches from execution feedback. (b) Identify behaviorally compatible skills through cross-instance replay and mechanism assessment. (c) Consolidate compatible skills only when the resulting abstraction preserves their validated behavior.}
\label{fig:skillcommit-process}
\end{figure*}

Reliable skill evolution requires generalizing individual experiences without compromising previously validated behavior.
\method achieves this through three stages of behaviorally validated scope expansion (\autoref{fig:skillcommit-process}).

\subsection{Feedback-Guided Patch Induction}
This stage aims to transform feedback from a failed task execution into a procedural patch that the frozen executor can use to improve its behavior.
It follows an iterative two-step process: generating a candidate patch from teacher feedback and validating the patch through source replay. %

\paragraph{Patch Generation.}
The core idea is to contrast the executor's failed behavior with a teacher-generated reference. %
Specifically, when the executor fails on a source instance $x_i$ and produces a trajectory $\tau_i^{-}$, a teacher model is provided with $x_i$, the verified ground-truth answer, and $\tau_i^{-}$.
Conditioned on this information, the teacher produces a reference trajectory $\tau_i^{+}$ that reaches the correct answer and diagnoses the discrepancy between the failed and successful behaviors.
Based on this contrast, the teacher generates a candidate patch $c_i$ containing concise procedural guidance, such as behavioral constraints, decision rules, or checkable intermediate steps intended to address the failure observed in $\tau_i^{-}$.
Note that the teacher may use the same underlying model as the executor, but operates under a separate teacher prompt with access to privileged feedback that is never exposed during execution.
To prevent solution leakage, candidates containing the final answer or an instance-specific solution to $x_i$ are discarded before validation.

\paragraph{Source-replay Validation.}
Each candidate patch $c_i$ is validated by replaying the source instance with the same frozen executor.
During replay, the executor receives only the original task context $x_i$ and the candidate patch $c_i$.
The resulting answer is evaluated by the task verifier, which retains access to the ground truth.
If the executor successfully solves $x_i$, the candidate is admitted as a source-validated instance-specific patch.
Otherwise, the rejected patch, the new replay trajectory, and the verifier feedback are returned to the skill author to guide the next generation attempt.
This generate-and-replay cycle stops at the first successful replay or after $K=3$ attempts.
In practice, \textit{88.0\%} of failed instances yield a validated patch. %

\subsection{Behavioral Compatibility Grouping}

This stage determines how a locally validated patch can be reused beyond its original instance.
Given an incoming patch $q_t$ and its source instance $x_t$, \method first examines whether $q_t$ is behaviorally compatible with an existing higher-level skill.
If a compatible skill is found, the framework attempts to expand that skill to cover $q_t$.
Otherwise, it searches for a group of compatible residual patches and uses them to induce a new higher-level skill.
In both cases, semantic similarity is used only to retrieve a bounded set of plausible candidates.
The actual decision is based on behavioral replay and mechanism assessment.
The prompt templates for compatibility assessment and higher-level skill induction are provided in Appendix~\ref{app:method-prompts}.

\paragraph{Expanding an Existing Skill.}
\method first retrieves the active higher-level skills whose descriptions are most similar to the incoming patch $q_t$.
We embed the repair text of $q_t$ and the structured description of each active skill, rank the active skill versions by cosine similarity, and retain the top $R$ candidates, with $R=4$ in our experiments.
For each retrieved skill $p$, \method applies $p$ and $q_t$ beyond their validated scopes by replaying $p$ on the incoming instance $x_t$ and $q_t$ on instances $\mathcal{W}(p)$ covered by $p$.
Let $\rho(m,x)\in{0,1}$ indicate whether the frozen executor solves instance $x$ when guided only by memory $m$.
We compute the behavioral compatibility between $p$ and $q_t$ as:

\begin{equation}
C(p,q_t)=\frac{
\rho(p,x_t)
+
\sum_{x_j\in\mathcal{W}(p)}
\rho(q_t,x_j)
}{
1+|\mathcal{W}(p)|
}.
\label{eq}
\end{equation}
The pair passes the behavioral compatibility screen when $C(p,q_t)\geq\gamma$, where $\gamma\in[0,1]$ is a predefined threshold.
This cross-replay provides graded evidence that the two memories operate over related behavioral contexts.

Because partial behavioral transfer may still arise from different strategies, a pattern author subsequently assesses whether $p$ and $q_t$ can be explained by a shared procedural mechanism.
If the compatibility score and mechanism assessment both support consolidation, the author proposes an expanded version of $p$ whose scope includes the incoming source.
The proposal remains a candidate until full-scope source replay confirms that it preserves validated behavior.

\paragraph{Forming a New Skill.}
If no existing skill passes the previous checks, \method proceeds to form a new skill from residual patches, i.e., those source-validated patches that have not yet been incorporated into any higher-level skill.
Similarly, \method retrieves a set of semantically related residual patches $\mathcal{C}_t$ and evaluates their behavioral compatibility through cross-instance replay.

Based on these replay outcomes, \method searches for an anchor patch whose procedure transfers to the source instances of all other patches in a candidate group. An eligible group must include the incoming patch $q_t$ and contain at least $m$ members.
By default, the anchor is required to transfer to every member, while reverse transfer may be imposed as a stricter optional criterion.
Among eligible groups, \method selects the one with the broadest behavioral coverage, using semantic similarity and a deterministic ordering only to resolve ties. If no eligible group is found, $q_t$ remains residual and no new skill is created.

\subsection{Source-Preserving Consolidation}
After identifying either a behaviorally coherent group of patches or a compatible existing skill, \method induces a candidate skill over the proposed scope.
Because abstraction may omit constraints essential to individual source instances, consolidation is governed by full source replay.

\paragraph{Candidate Skill Induction.}
A candidate skill can arise in two ways.
If the previous stage identifies a compatible group of residual patches, the pattern author attempts to induce a new higher-level skill from that group.
If the incoming patch is compatible with an existing skill, the pattern author produces an expanded version of that skill.
In both cases, the induced skill specifies an applicability condition, executable procedural guidance, and an exclusion condition, represented by \texttt{when}, \texttt{repair}, and \texttt{avoid\_when}, respectively.
To encourage mechanism-level abstraction rather than instance memorization, the pattern author is not given source questions, answers, teacher trajectories, or provenance information.
For forming a new skill, it receives only the anonymized patch texts in the proposed group.
For an expansion, it additionally receives the current skill description.

\paragraph{Full-Scope Source Replay.}
Let $\mathcal{S}$ denote the proposed source set of a candidate.
To verify that the abstraction preserves previously validated behavior, \method replays the candidate on every source instance in $\mathcal{S}$ using the frozen executor.
If replay succeeds on every source, the candidate is regarded as fully supported.
Otherwise, the pattern author incrementally refines the current candidate using its previous version together with the replay trajectories and verifier feedback, while ground-truth answers and instance identities remain hidden.
Each refinement produces a new candidate version, which is replayed on the same source set before further refinement.
This process continues for at most $L$ attempts, and all candidate versions together with their replay outcomes are retained for auditing.

\paragraph{Source-Preserving Commit and Split.}
To ensure that every committed skill preserves the validated behavior of every source instance within its declared scope, \method uses source replay as the sole criterion for consolidation. Depending on the replay outcome, each candidate skill follows one of three transitions.

\textbf{Commit.}
A fully supported candidate is committed as a higher-level skill.
When the candidate expands an existing skill, the new version replaces the active version while preserving previous versions for traceability.
When the candidate is induced from a group of residual patches, it becomes a new active skill.

\textbf{Split.}
If replay succeeds on only a subset of the proposed sources, \method selects the candidate version with the strongest replay support.
If the supported source set of this version contains at least $m$ instances, it will be split by restricting its scope to the supported sources and replayed once more on this reduced scope.
A successfully revalidated split is committed as an independent higher-level skill.
Typically, we set $m\in\{2,3\}$ depending on the dataset (Appendix~\ref{app:hyperparameters-budgets}).

\textbf{Reject.}
If no candidate the candidate is discarded. The existing skill hierarchy remains unchanged, and the corresponding patches continue to be retained as instance-specific skills.
If no candidate version yields a replay-supported subset of at least $m$ source instances, or if the narrowed candidate fails revalidation, the candidate is rejected. The existing skill hierarchy remains unchanged, while the corresponding patches remain available as instance-specific skills.

\section{Experiments}
\begin{table*}[!t]
\centering
\normalfont\footnotesize
\setlength{\tabcolsep}{4pt}
\renewcommand{\arraystretch}{0.96}
\setlength{\aboverulesep}{1.2pt}
\setlength{\belowrulesep}{1.6pt}
\begin{tabular*}{\textwidth}{@{\extracolsep{\fill}}lrrrrrrr@{}}
\toprule
\textbf{Method} & \multicolumn{2}{c}{\textbf{RuleArena}} & \multicolumn{2}{c}{\textbf{OpenExempt}} & \multicolumn{2}{c}{\textbf{KOR-Bench}} & \textbf{Avg.}$\uparrow$ \\
\cmidrule(lr){2-3}\cmidrule(lr){4-5}\cmidrule(lr){6-7}
& \textbf{Airline}$\uparrow$ & \textbf{NBA}$\uparrow$ & \textbf{EV}$\uparrow$ & \textbf{EC}$\uparrow$ & \textbf{Logic}$\uparrow$ & \textbf{Cipher}$\uparrow$ & \\
\midrule
\rowcolor{black!5}
\multicolumn{8}{l}{\textbf{Skill author:} \texttt{DeepSeek-V4-Flash} \quad \textbf{Skill user:} \texttt{DeepSeek-V4-Flash}} \\
\addlinespace[0.5pt]
No Skill & \cellcolor{SCtableblue!12}72.00 & \cellcolor{SCtableblue!10}28.89 & \cellcolor{SCtableblue!14}76.86 & \cellcolor{SCtableblue!14}86.33 & \cellcolor{SCtableblue!10}72.00 & \cellcolor{SCtableblue!6}73.33 & 68.24 \\
ExpeL & \cellcolor{SCtableblue!19}88.00 & \cellcolor{SCtableblue!17}53.33 & \cellcolor{SCtableblue!3}50.17 & \cellcolor{SCtableblue!15}88.36 & \cellcolor{SCtableblue!19}\underline{80.00} & \cellcolor{SCtableblue!13}78.67 & 73.09 \\
ReasoningBank & \cellcolor{SCtableblue!10}66.00 & \cellcolor{SCtableblue!10}31.11 & \cellcolor{SCtableblue!10}65.54 & \cellcolor{SCtableblue!3}74.25 & \cellcolor{SCtableblue!6}68.00 & \cellcolor{SCtableblue!5}72.00 & 62.82 \\
ACE & \cellcolor{SCtableblue!3}51.00 & \cellcolor{SCtableblue!3}6.67 & \cellcolor{SCtableblue!19}\underline{87.32} & \cellcolor{SCtableblue!22}\underline{95.65} & \cellcolor{SCtableblue!10}72.00 & \cellcolor{SCtableblue!3}70.67 & 63.89 \\
Trace2Skill & \cellcolor{SCtableblue!3}50.00 & \cellcolor{SCtableblue!10}28.89 & \cellcolor{SCtableblue!18}84.30 & \cellcolor{SCtableblue!4}75.50 & \cellcolor{SCtableblue!9}70.67 & \cellcolor{SCtableblue!3}70.67 & 63.34 \\
SkillCAT & \cellcolor{SCtableblue!8}61.00 & \cellcolor{SCtableblue!19}\underline{60.00} & \cellcolor{SCtableblue!13}73.24 & \cellcolor{SCtableblue!10}82.54 & \cellcolor{SCtableblue!9}70.67 & \cellcolor{SCtableblue!11}77.33 & 70.80 \\
GSE & \cellcolor{SCtableblue!20}\underline{92.00} & \cellcolor{SCtableblue!12}37.78 & \cellcolor{SCtableblue!15}78.40 & \cellcolor{SCtableblue!11}83.50 & \cellcolor{SCtableblue!3}65.33 & \cellcolor{SCtableblue!20}\underline{85.33} & \underline{73.72} \\
\addlinespace[1.5pt]
\textbf{\method{} (Ours)} & \cellcolor{SCtableblue!22}\textbf{96.00} & \cellcolor{SCtableblue!22}\textbf{71.11} & \cellcolor{SCtableblue!22}\textbf{94.28} & \cellcolor{SCtableblue!22}\textbf{96.01} & \cellcolor{SCtableblue!22}\textbf{82.67} & \cellcolor{SCtableblue!22}\textbf{86.67} & \textbf{87.79} \\
\midrule
\rowcolor{black!5}
\multicolumn{8}{l}{\textbf{Skill author:} \texttt{Qwen3.6-35B-A3B} \quad \textbf{Skill user:} \texttt{Qwen3.6-35B-A3B}} \\
\addlinespace[0.5pt]
No Skill & \cellcolor{SCtableblue!10}51.00 & \cellcolor{SCtableblue!8}26.67 & \cellcolor{SCtableblue!9}29.02 & \cellcolor{SCtableblue!12}\underline{47.46} & \cellcolor{SCtableblue!3}53.33 & \cellcolor{SCtableblue!7}76.00 & 47.25 \\
ExpeL & \cellcolor{SCtableblue!11}54.00 & \cellcolor{SCtableblue!10}31.11 & \cellcolor{SCtableblue!3}7.76 & \cellcolor{SCtableblue!3}21.53 & \cellcolor{SCtableblue!14}65.33 & \cellcolor{SCtableblue!11}78.67 & 43.07 \\
ReasoningBank & \cellcolor{SCtableblue!7}44.00 & \cellcolor{SCtableblue!8}26.67 & \cellcolor{SCtableblue!7}22.37 & \cellcolor{SCtableblue!5}27.64 & \cellcolor{SCtableblue!8}58.67 & \cellcolor{SCtableblue!5}74.67 & 42.34 \\
ACE & \cellcolor{SCtableblue!3}32.00 & \cellcolor{SCtableblue!3}11.11 & \cellcolor{SCtableblue!12}\underline{38.42} & \cellcolor{SCtableblue!11}45.71 & \cellcolor{SCtableblue!21}\underline{73.33} & \cellcolor{SCtableblue!9}77.33 & 46.32 \\
Trace2Skill & \cellcolor{SCtableblue!11}\underline{55.00} & \cellcolor{SCtableblue!10}31.11 & \cellcolor{SCtableblue!11}35.68 & \cellcolor{SCtableblue!6}29.34 & \cellcolor{SCtableblue!10}61.33 & \cellcolor{SCtableblue!18}\underline{82.67} & \underline{49.19} \\
SkillCAT & \cellcolor{SCtableblue!9}49.00 & \cellcolor{SCtableblue!8}26.67 & \cellcolor{SCtableblue!7}22.37 & \cellcolor{SCtableblue!4}24.49 & \cellcolor{SCtableblue!4}54.67 & \cellcolor{SCtableblue!3}73.33 & 41.76 \\
GSE & \cellcolor{SCtableblue!9}48.00 & \cellcolor{SCtableblue!11}\underline{33.33} & \cellcolor{SCtableblue!11}35.99 & \cellcolor{SCtableblue!8}35.76 & \cellcolor{SCtableblue!9}60.00 & \cellcolor{SCtableblue!9}77.33 & 48.40 \\
\addlinespace[1.5pt]
\textbf{\method{} (Ours)} & \cellcolor{SCtableblue!22}\textbf{85.00} & \cellcolor{SCtableblue!22}\textbf{66.67} & \cellcolor{SCtableblue!22}\textbf{72.46} & \cellcolor{SCtableblue!22}\textbf{79.46} & \cellcolor{SCtableblue!22}\textbf{74.67} & \cellcolor{SCtableblue!22}\textbf{85.33} & \textbf{77.27} \\
\midrule
\rowcolor{black!5}
\multicolumn{8}{l}{\textbf{Skill author:} \texttt{Qwen3.6-35B-A3B} \quad \textbf{Skill user:} \texttt{Qwen3.5-9B}} \\
\addlinespace[0.5pt]
No Skill & \cellcolor{SCtableblue!15}55.00 & \cellcolor{SCtableblue!3}15.56 & \cellcolor{SCtableblue!5}9.97 & \cellcolor{SCtableblue!5}15.61 & \cellcolor{SCtableblue!13}60.00 & \cellcolor{SCtableblue!17}65.33 & 36.91 \\
ExpeL & \cellcolor{SCtableblue!14}54.00 & \cellcolor{SCtableblue!8}26.67 & \cellcolor{SCtableblue!3}4.18 & \cellcolor{SCtableblue!3}10.73 & \cellcolor{SCtableblue!15}\underline{62.67} & \cellcolor{SCtableblue!18}70.67 & 38.15 \\
ReasoningBank & \cellcolor{SCtableblue!14}53.00 & \cellcolor{SCtableblue!6}22.22 & \cellcolor{SCtableblue!8}16.42 & \cellcolor{SCtableblue!7}22.68 & \cellcolor{SCtableblue!9}52.00 & \cellcolor{SCtableblue!17}65.33 & 38.61 \\
ACE & \cellcolor{SCtableblue!3}14.00 & \cellcolor{SCtableblue!3}15.56 & \cellcolor{SCtableblue!11}26.31 & \cellcolor{SCtableblue!12}34.57 & \cellcolor{SCtableblue!3}41.33 & \cellcolor{SCtableblue!19}\underline{74.67} & 34.41 \\
Trace2Skill & \cellcolor{SCtableblue!9}35.00 & \cellcolor{SCtableblue!10}\underline{31.11} & \cellcolor{SCtableblue!11}24.76 & \cellcolor{SCtableblue!8}25.43 & \cellcolor{SCtableblue!13}60.00 & \cellcolor{SCtableblue!19}73.33 & 41.61 \\
SkillCAT & \cellcolor{SCtableblue!15}\underline{56.00} & \cellcolor{SCtableblue!7}24.44 & \cellcolor{SCtableblue!10}21.83 & \cellcolor{SCtableblue!12}\underline{35.99} & \cellcolor{SCtableblue!13}58.67 & \cellcolor{SCtableblue!3}18.67 & 35.93 \\
GSE & \cellcolor{SCtableblue!15}55.00 & \cellcolor{SCtableblue!8}26.67 & \cellcolor{SCtableblue!14}\underline{32.18} & \cellcolor{SCtableblue!6}18.86 & \cellcolor{SCtableblue!10}54.67 & \cellcolor{SCtableblue!19}72.00 & \underline{43.23} \\
\addlinespace[1.5pt]
\textbf{\method{} (Ours)} & \cellcolor{SCtableblue!22}\textbf{81.00} & \cellcolor{SCtableblue!22}\textbf{60.00} & \cellcolor{SCtableblue!22}\textbf{54.28} & \cellcolor{SCtableblue!22}\textbf{62.41} & \cellcolor{SCtableblue!22}\textbf{76.00} & \cellcolor{SCtableblue!22}\textbf{84.00} & \textbf{69.62} \\
\bottomrule
\end{tabular*}
\caption{\textbf{Main results (\%).} RuleArena and KOR-Bench report strict accuracy; OpenExempt reports macro-F1.
Bold/underline mark best/second-best scores per configuration (including ties); darker blue indicates higher scores within each task/configuration. Avg. is the unweighted mean of the six displayed task scores.}
\label{tab:main-results}
\end{table*}

\providecommand{\AppendixRef}[1]{\begingroup
    \def\appendixautorefname{Appendix}%
    \def\sectionautorefname{Appendix}%
    \def\subsectionautorefname{Appendix}%
    \autoref{#1}\endgroup}

We evaluate \method through three research questions: whether it improves task performance (RQ1, \S\,\ref{sec:rq-effectiveness}), how its components contribute (RQ2, \S\,\ref{sec:rq-components}), and whether learned skills transfer across model scales and families (RQ3, \S\,\ref{sec:rq-transfer}).

\subsection{Experimental Setup}

\paragraph{Benchmarks.}
We evaluate six tasks across three domains: rule-based reasoning with RuleArena's Airline and NBA \citep{zhou2025rulearena}; legal reasoning with Exemption Classification (EC) and Exemption Valuation (EV) from OpenExempt's Advanced Competency suite \citep{servantez2026openexempt}; and knowledge-orthogonal abstract reasoning with KOR-Bench's Logic and Cipher \citep{ma2025korbench}.
This task suite tests whether experience-derived guidance remains useful across different reasoning requirements beyond one single evaluation environment.
Data splits and evaluation interfaces are detailed in \AppendixRef{app:evaluation-interfaces}.

\paragraph{Baselines.}
We compare with No Skill and six skill evolution and experience learning methods.
ExpeL \citep{zhao2024expel} extracts reusable insights, ReasoningBank \citep{ouyang2026reasoningbank} retrieves trajectory memories, ACE \citep{zhang2026ace} evolves an external playbook, and Trace2Skill \citep{ni2026trace2skill} hierarchically distills trajectory-level lessons into a skill document.
Two closely related skill-evolution methods are SkillCAT \citep{chen2026skillcat}, which validates source-task patches before hierarchical merging, and GSE \citep{yang2026gse}, which combines relation-aware evolution, cluster consolidation, and historical replay.
These comparisons situate \method among both memory-construction methods and methods that also validate skill evolution.
Implementations and shared evaluation settings are detailed in \AppendixRef{app:baseline-implementations}.

\paragraph{LLM Backbones.}
We separate matched-model effectiveness from transfer across model scales and families.
\texttt{DeepSeek-V4-Flash} and \texttt{Qwen3.6-35B-A3B} each serve as both skill author and skill user in the matched settings.
We then deploy \texttt{Qwen3.6-35B-A3B}-authored skills with the smaller \texttt{Qwen3.5-9B} user to examine in-family, cross-scale transfer.
Using \texttt{GPT-5.4} as the author and the same \texttt{Qwen3.5-9B} user further examines cross-family transfer.
All model parameters remain frozen throughout skill construction and evaluation, with generation settings and method-specific budgets provided in \AppendixRef{app:experimental-details}.

\paragraph{Evaluation Metrics.}
We report strict accuracy for RuleArena and KOR-Bench and macro-F1 using OpenExempt's official evaluator.
On Airline and NBA with \texttt{DeepSeek-V4-Flash}, we also decompose prediction changes into Help and Hurt.
Help denotes corrected errors under No Skill to correct with a method; Hurt denotes newly introduced errors.

\subsection{Does \method Improve Accuracy?}
\label{sec:rq-effectiveness}

\textbf{\method consistently improves task performance across domains and model configurations.}
As shown in \autoref{tab:main-results}, \method outperforms No Skill and all six baselines across the six tasks and three model configurations.
Its average task score exceeds the strongest baseline in each configuration by \textbf{14.07}, \textbf{28.08}, and \textbf{26.39} points, respectively.
These results show that the learned guidance improves downstream decisions across different reasoning domains and model configurations, extending its utility beyond a single task or executor.

\begin{figure}[t]
\centering
\includegraphics[width=\columnwidth]{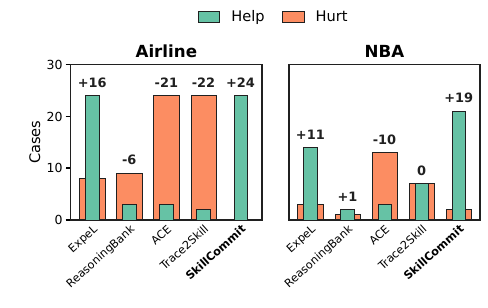}
\vspace{-10mm}
\caption{
Error corrections and regressions on RuleArena.
Help denotes corrected errors, Hurt denotes newly introduced errors, and labels indicate net improvements (Help $-$ Hurt).}
\label{fig:rulearena-help-hurt}
\end{figure}

\textbf{Performance gains combine error correction with preservation of correct decisions.}
On Airline, \method and ExpeL both repair \textbf{24} failures, but ExpeL introduces \textbf{8} regressions while \method introduces \textbf{0} (\autoref{fig:rulearena-help-hurt}).
The eight-point accuracy gap therefore comes entirely from avoiding regressions, not from repairing more failures.
On NBA, \method repairs 21 failures while introducing two regressions, yielding \textbf{19} additional correct decisions.
Together, these comparisons show why repair counts alone are insufficient to assess skill utility: useful guidance must also preserve decisions the executor already gets right.
The following ablations examine which construction stages contribute to these gains.

\subsection{Why Does \method Work?}
\label{sec:rq-components}

\autoref{tab:ablation} compares stage removal and isolation on NBA and EC with \texttt{DeepSeek-V4-Flash}.
Without patch induction, first-pass teacher patches remain but validation and refinement are removed; without consolidation, synthesis remains but replay verification is removed.

\begin{table}[t]
\centering
{\footnotesize
\newcommand{\ablationDrop}[2]{$#1_{\makebox[31pt][l]{\scriptsize\textcolor[rgb]{0.2,0.2,1}{$(-#2\%)$}}}$}
\newcommand{\ablationRef}[1]{$#1_{\makebox[31pt]{}}$}
\setlength{\tabcolsep}{3pt}
\renewcommand{\arraystretch}{1.12}
\begin{tabular*}{\linewidth}{@{\extracolsep{\fill}}lrr@{}}
\toprule
\textbf{Variant} & \textbf{NBA} (Acc.$\uparrow$) & \textbf{EC} (Macro-F1$\uparrow$) \\
\midrule
\textbf{Full} & \ablationRef{71.11} & \ablationRef{96.01} \\
No Skill & \ablationDrop{28.89}{59.4} & \ablationDrop{86.33}{10.1} \\
\midrule
\rowcolor{black!15}
\multicolumn{3}{l}{\textbf{Stage removal}} \\
\textit{w/o} Patch Induction & \ablationDrop{53.33}{25.0} & \ablationDrop{91.41}{4.8} \\
\textit{w/o} Grouping & \ablationDrop{42.22}{40.6} & \ablationDrop{88.87}{7.4} \\
\textit{w/o} Consolidation & \ablationDrop{48.89}{31.3} & \ablationDrop{89.12}{7.2} \\
\midrule
\rowcolor{black!15}
\multicolumn{3}{l}{\textbf{Stage isolation}} \\
Patch Induction \textit{only} & \ablationDrop{33.33}{53.1} & \ablationDrop{86.97}{9.4} \\
Grouping \textit{only} & \ablationDrop{40.00}{43.8} & \ablationDrop{88.54}{7.8} \\
Consolidation \textit{only} & \ablationDrop{35.56}{50.0} & \ablationDrop{87.59}{8.8} \\
\bottomrule
\end{tabular*}
}
\caption{\textbf{Ablation study of \method{}.}
Subscripts show relative changes from Full.}
\label{tab:ablation}
\end{table}

\textbf{Validated source patches provide a stronger basis for skill construction.}
Removing patch validation and refinement lowers NBA accuracy by \textbf{17.78} points and EC macro-F1 by \textbf{4.60} points relative to Full.
Although grouping and consolidation remain, they do not recover the performance obtained from validated patches.
This supports checking the reliability of local repairs before using them as inputs to broader skill construction.

\textbf{Behavioral grouping and consolidation verification address distinct risks in abstraction.}
Removing grouping produces the largest observed drop on both tasks, reducing correct NBA decisions from \textbf{32 to 19} and EC macro-F1 by \textbf{7.14} points.
Qualitative inspection of NBA skills found that semantically similar patches could encode conflicting mechanisms; merging them omitted conditions essential to the original repairs.
However, compatible inputs alone do not ensure a faithful synthesis: removing consolidation verification leaves only \textbf{22} correct NBA decisions and reduces EC macro-F1 by \textbf{6.89} points.
These results support both selecting behaviorally compatible inputs and replaying the resulting abstraction to check that their useful behavior survives consolidation.

\textbf{No isolated stage recovers the gains of the full pipeline.}
The isolated stages add only \textbf{2--5} correct NBA decisions over No Skill, compared with \textbf{19} for Full; on EC, their gains are \textbf{0.64--2.21} points versus \textbf{9.68}.
Inspection of the isolated-stage outputs helps explain this gap: validated patches can retain instance-specific constraints that limit reuse, while consolidation of heterogeneous, unvalidated patches can lose useful repair mechanisms.
Together, the removal and isolation results support combining reliable local repairs, compatible grouping, and verified abstraction rather than relying on any single stage.
Library construction outcomes and costs are reported in \AppendixRef{app:additional-results}.

\begin{figure}[t]
\centering
\includegraphics[width=\columnwidth]{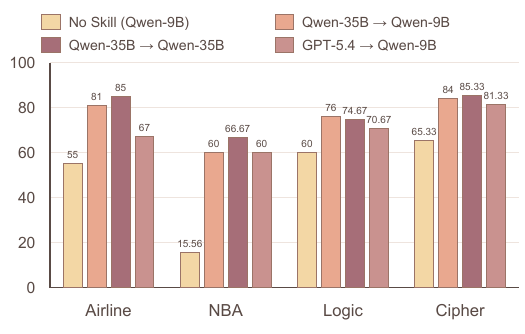}
\vspace{-8mm}
\caption{
Skill transfer across model scales and families. Arrows indicate skill author $\rightarrow$ frozen skill user.
}
\label{fig:cross-model-transfer}
\end{figure}

\subsection{Can Skills Be Reused by Other Models?}
\label{sec:rq-transfer}

\autoref{fig:cross-model-transfer} evaluates skills from two author families with the same \texttt{Qwen3.5-9B} executor. %

\textbf{Learned skills improve decisions beyond the author model.}
Skills authored by \texttt{Qwen3.6-35B-A3B} and \texttt{GPT-5.4} improve \textbf{all four tasks} over No Skill, with average accuracy gains of \textbf{26.28} and \textbf{20.78} percentage points, respectively.
The former demonstrates reuse by a smaller executor within the same family; the latter extends this benefit across model families.
Thus, the learned guidance remains useful when the model producing the skills differs from the model applying them.
The Qwen-authored library leads by \textbf{5.50} points on average, although both libraries reach \textbf{60.00\%} on NBA.
This highlights the author--executor pairing as a practical consideration when constructing a reusable skill library.

\providecommand{\CaseStudyText}[1]{#1}

\subsection{\texorpdfstring{\CaseStudyText{Case Study}}{Case Study}}
\label{sec:case-study}

\CaseStudyText{\textbf{Replay determines how far a local repair can be generalized.}
\autoref{fig:skill-evolution-case} illustrates this process with $P_{24}$, an asset-exclusion patch.
When retrieved skills fail the compatibility check, residual replay groups it with $P_{19}$, $P_{20}$, and $P_{21}$, while excluding $x_{22}$.
The consolidation panels then illustrate alternative outcomes: full-source support permits a new skill or parent replacement, whereas partial support requires a narrower abstraction.}

\CaseStudyText{\textbf{Narrowing retains useful behavior without forcing unsupported coverage.}
In panel (e), revalidation supports sources 19, 20, and 24, while patch 21 remains residual.
Panel (f) instead commits a sibling covering sources 5--7 and 24 while preserving $S_2$ over 5--10.
These paths retain the supported repair without discarding the parent's existing coverage; failed candidates leave both patches and parents unchanged.}

\CaseStudyText{\textbf{The logged lineage shows expansion alternating with rejected proposals.}
\autoref{fig:case-study-multi-skill-lineages} tracks seven selected EC skill families across 126 repair arrivals, comprising 17 committed versions and 17 uncommitted candidates.
The eligibility skill $S_B$ grows from \textbf{two to five sources} through four versions.
At arrival 102, two candidate revisions each pass only \textbf{4/5} source replays before a subsequent revision is committed with full support.
At arrival 110, three attempts to add a sixth source fail full-source replay, so the five-source version remains active.
This sequence shows how verification both enables supported expansion and prevents an unsuccessful update from replacing an established skill.}

\providecommand{\CaseStudyText}[1]{#1}

\begin{figure}[!t]
\centering
\includegraphics[width=\columnwidth]{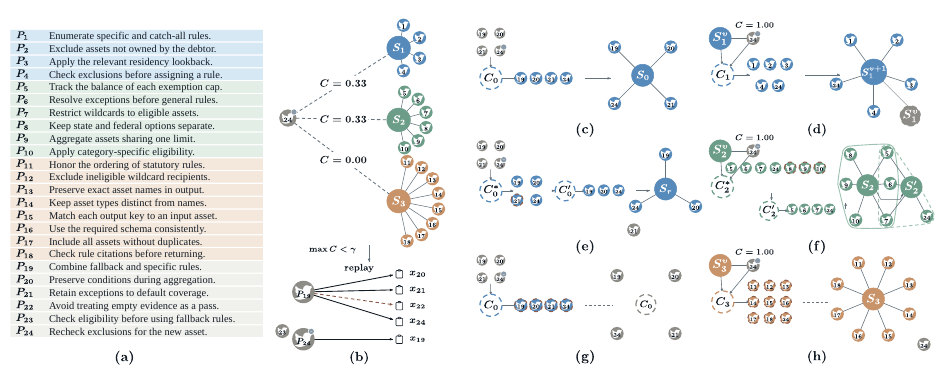}
\caption{\CaseStudyText{Illustrative patch evolution in exemption reasoning:
(a) local patches; (b) retrieval and replay; (c,e,g) new-skill consolidation; (d,f,h) skill expansion.
Rows show full support, partial support, and rejection; brown marks failed checks.}}
\label{fig:skill-evolution-case}
\end{figure}

\begin{figure}[!t]
\centering
\includegraphics[width=\columnwidth]{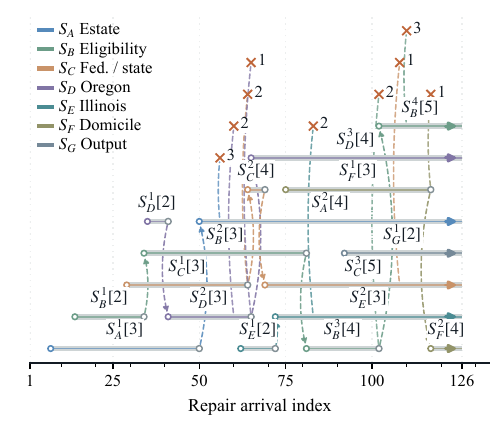}
\vspace{-8mm}
\caption{\CaseStudyText{Skill evolution on EC.
Bars show version lifespans, dashed links show parentage, and arrows mark final active versions.
Crosses mark uncommitted candidates with counts.
$S_i^v[n]$ labels family, version, and source count.}}
\label{fig:case-study-multi-skill-lineages}
\end{figure}

\section{Conclusion}
We presented \method, an online skill evolution framework that transforms accumulated agent experience into a compact, hierarchical skill library without sacrificing previously validated behavior.
Rather than relying solely on semantic similarity or model judgment, \method admits every broader behavioral claim only after it survives replay.
Experiments demonstrate consistent improvements over no-skill settings and strong skill-evolution baselines.
The resulting skills also transfer across model scales and families, suggesting that skills can be authored by more capable models and reused by more economical agents.
\FloatBarrier
\section*{Limitations}
The evaluation uses benchmark tasks with ground-truth verifiers, so the current
evidence does not establish effectiveness in open-ended environments without
reliable verification. Replay checks behavior on observed source instances;
they do not certify every input within a skill's stated applicability region.
The stage-level ablation is limited to RuleArena NBA and OpenExempt EC, and the paper does not
report run-to-run variance across the full task suite. It also does not report
library growth across the full induction stream or a matched total-induction
cost comparison against every baseline.

\section*{Ethical Considerations}
\method studies skill construction and reuse on existing reasoning benchmarks.
Its replay checks assess task correctness on observed instances; they do not
establish that a skill is safe, fair, or appropriate for every future use.
A library may preserve or amplify errors and biases inherited from its source
trajectories, underlying models, or task verifiers. Retaining source evidence
and skill versions can support auditing, but does not replace independent
safety and fairness evaluation.

The legal reasoning tasks in our evaluation are research benchmarks, and their
scores should not be interpreted as evidence of suitability for legal advice
or autonomous decisions affecting individuals. Such applications require
qualified human oversight and validation in the intended setting. When
extending the method to user trajectories, practitioners should obtain
appropriate authorization, minimize retained personal information, and review
what is sent to external model providers or included in shared skill libraries.
Data and model use must respect applicable licenses and access terms.
Extensions that grant agents tool or code execution privileges should use
isolated environments, restricted permissions, and human approval for
consequential actions. Behavioral replay alone is not a security boundary.

\bibliography{references}

\begin{thebibliography}{30}
\providecommand{\natexlab}[1]{#1}

\bibitem[{Chen et~al.(2026)Chen, Zhong, Liu, and Du}]{chen2026skillcat}
Kunfeng Chen, Qihuang Zhong, Juhua Liu, and Bo~Du. 2026.
\newblock \href {https://arxiv.org/abs/2606.13317} {{SkillCAT}: Contrastive,
  assessment-augmented and topology-aware skill self-evolution for {LLM}
  agents}.
\newblock \emph{Preprint}, arXiv:2606.13317.

\bibitem[{Chi et~al.(1981)Chi, Feltovich, and Glaser}]{chi1981categorization}
Michelene T.~H. Chi, Paul~J. Feltovich, and Robert Glaser. 1981.
\newblock \href {https://doi.org/10.1207/s15516709cog0502_2} {Categorization
  and representation of physics problems by experts and novices}.
\newblock \emph{Cognitive Science}, 5(2):121--152.

\bibitem[{Deng et~al.(2023)Deng, Gu, Zheng, Chen, Stevens, Wang, Sun, and
  Su}]{deng2023mind2web}
Xiang Deng, Yu~Gu, Boyuan Zheng, Shijie Chen, Sam Stevens, Boshi Wang, Huan
  Sun, and Yu~Su. 2023.
\newblock \href
  {https://proceedings.neurips.cc/paper_files/paper/2023/hash/5950bf290a1570ea401bf98882128160-Abstract-Datasets_and_Benchmarks.html}
  {{Mind2Web}: Towards a generalist agent for the web}.
\newblock In \emph{Advances in Neural Information Processing Systems},
  volume~36, pages 28091--28114. Curran Associates, Inc.

\bibitem[{Gentner(1983)}]{gentner1983structure}
Dedre Gentner. 1983.
\newblock \href {https://doi.org/10.1207/s15516709cog0702_3}
  {Structure-mapping: A theoretical framework for analogy}.
\newblock \emph{Cognitive Science}, 7(2):155--170.

\bibitem[{Gick and Holyoak(1983)}]{gick1983schema}
Mary~L. Gick and Keith~J. Holyoak. 1983.
\newblock \href {https://doi.org/10.1016/0010-0285(83)90002-6} {Schema
  induction and analogical transfer}.
\newblock \emph{Cognitive Psychology}, 15(1):1--38.

\bibitem[{Gupta et~al.(2024)Gupta, Kirtania, Singha, Gulwani, Radhakrishna,
  Soares, and Shi}]{gupta2024metareflection}
Priyanshu Gupta, Shashank Kirtania, Ananya Singha, Sumit Gulwani, Arjun
  Radhakrishna, Gustavo Soares, and Sherry Shi. 2024.
\newblock \href {https://doi.org/10.18653/v1/2024.emnlp-main.477}
  {{MetaReflection}: Learning instructions for language agents using past
  reflections}.
\newblock In \emph{Proceedings of the 2024 Conference on Empirical Methods in
  Natural Language Processing}, pages 8369--8385. Association for Computational
  Linguistics.

\bibitem[{Kolb(2014)}]{kolb2014experiential}
David~A Kolb. 2014.
\newblock \emph{Experiential learning: Experience as the source of learning and
  development}.
\newblock FT press.

\bibitem[{Li et~al.(2026)Li, Liu, Chen, You, Di, He, Zheng
  et~al.}]{li2026skillsbench}
Xiangyi Li, Yimin Liu, Wenbo Chen, Bingran You, Zonglin Di, Yifeng He, Shenghan
  Zheng, et~al. 2026.
\newblock \href {https://arxiv.org/abs/2602.12670} {{SkillsBench}: Benchmarking
  how well agent skills work across diverse tasks}.
\newblock \emph{Preprint}, arXiv:2602.12670.

\bibitem[{Ma et~al.(2025)Ma, Du, Wang, Zhang, Wen, Qu, Yang, Liu, Liu, Yue,
  Huang, and Zhang}]{ma2025korbench}
Kaijing Ma, Xinrun Du, Yunran Wang, Haoran Zhang, Zhoufutu Wen, Xingwei Qu,
  Jian Yang, Jiaheng Liu, Minghao Liu, Xiang Yue, Wenhao Huang, and Ge~Zhang.
  2025.
\newblock \href {https://openreview.net/forum?id=SVRRQ8goQo} {{KOR-Bench}:
  Benchmarking language models on knowledge-orthogonal reasoning tasks}.
\newblock In \emph{The Thirteenth International Conference on Learning
  Representations}.

\bibitem[{Ma et~al.(2026)Ma, Huang, Bao, Zhuang, Shukla, Galley, Zhang, and
  Feuerriegel}]{ma2026skillgen}
Yuchen Ma, Yue Huang, Han Bao, Haomin Zhuang, Swadheen Shukla, Michel Galley,
  Xiangliang Zhang, and Stefan Feuerriegel. 2026.
\newblock \href {https://arxiv.org/abs/2605.10999} {{SkillGen}: Verified
  inference-time agent skill synthesis}.
\newblock \emph{Preprint}, arXiv:2605.10999.

\bibitem[{McClelland et~al.(1995)McClelland, McNaughton, and
  O'Reilly}]{mcclelland1995complementary}
James~L. McClelland, Bruce~L. McNaughton, and Randall~C. O'Reilly. 1995.
\newblock \href {https://doi.org/10.1037/0033-295X.102.3.419} {Why there are
  complementary learning systems in the hippocampus and neocortex: Insights
  from the successes and failures of connectionist models of learning and
  memory}.
\newblock \emph{Psychological Review}, 102(3):419--457.

\bibitem[{Ni et~al.(2026)Ni, Liu, Liu, Sun, Zhou, Cheng, Wang, Zhao, Jiang, and
  Jiang}]{ni2026trace2skill}
Jingwei Ni, Yihao Liu, Xinpeng Liu, Yutao Sun, Mengyu Zhou, Pengyu Cheng, Dexin
  Wang, Erchao Zhao, Xiaoxi Jiang, and Guanjun Jiang. 2026.
\newblock \href {https://arxiv.org/abs/2603.25158} {{Trace2Skill}: Distill
  trajectory-local lessons into transferable agent skills}.
\newblock \emph{Preprint}, arXiv:2603.25158.

\bibitem[{Ouyang et~al.(2026)Ouyang, Yan, Hsu, Chen, Jiang, Wang, Han, Le,
  Daruki, Tang, Tirumalashetty, Lee, Rofouei, Lin, Han, Lee, and
  Pfister}]{ouyang2026reasoningbank}
Siru Ouyang, Jun Yan, I-Hung Hsu, Yanfei Chen, Ke~Jiang, Zifeng Wang, Rujun
  Han, Long~T. Le, Samira Daruki, Xiangru Tang, Vishy Tirumalashetty, George
  Lee, Mahsan Rofouei, Hangfei Lin, Jiawei Han, Chen-Yu Lee, and Tomas Pfister.
  2026.
\newblock \href {https://openreview.net/forum?id=jL7fwchScm} {{ReasoningBank}:
  Scaling agent self-evolving with reasoning memory}.
\newblock In \emph{The Fourteenth International Conference on Learning
  Representations}.

\bibitem[{Schick et~al.(2023)Schick, Dwivedi-Yu, Dessi, Raileanu, Lomeli,
  Hambro, Zettlemoyer, Cancedda, and Scialom}]{schick2023toolformer}
Timo Schick, Jane Dwivedi-Yu, Roberto Dessi, Roberta Raileanu, Maria Lomeli,
  Eric Hambro, Luke Zettlemoyer, Nicola Cancedda, and Thomas Scialom. 2023.
\newblock \href
  {https://proceedings.neurips.cc/paper_files/paper/2023/hash/d842425e4bf79ba039352da0f658a906-Abstract-Conference.html}
  {Toolformer: Language models can teach themselves to use tools}.
\newblock In \emph{Advances in Neural Information Processing Systems},
  volume~36, pages 68539--68551. Curran Associates, Inc.

\bibitem[{Servantez et~al.(2026)Servantez, Lawsky, Jain, Linna, and
  Hammond}]{servantez2026openexempt}
Sergio Servantez, Sarah~B. Lawsky, Rajiv Jain, Daniel~W. Linna, and Kristian
  Hammond. 2026.
\newblock \href {https://arxiv.org/abs/2601.13183} {{OpenExempt}: A diagnostic
  benchmark for legal reasoning and a framework for creating custom benchmarks
  on demand}.
\newblock \emph{Preprint}, arXiv:2601.13183.

\bibitem[{Shinn et~al.(2023)Shinn, Cassano, Gopinath, Narasimhan, and
  Yao}]{shinn2023reflexion}
Noah Shinn, Federico Cassano, Ashwin Gopinath, Karthik Narasimhan, and Shunyu
  Yao. 2023.
\newblock \href
  {https://proceedings.neurips.cc/paper_files/paper/2023/hash/1b44b878bb782e6954cd888628510e90-Abstract-Conference.html}
  {Reflexion: Language agents with verbal reinforcement learning}.
\newblock In \emph{Advances in Neural Information Processing Systems},
  volume~36.

\bibitem[{Suzgun et~al.(2026)Suzgun, Yuksekgonul, Bianchi, Jurafsky, and
  Zou}]{suzgun2026dynamic}
Mirac Suzgun, Mert Yuksekgonul, Federico Bianchi, Dan Jurafsky, and James Zou.
  2026.
\newblock \href {https://doi.org/10.18653/v1/2026.eacl-long.333} {Dynamic
  cheatsheet: Test-time learning with adaptive memory}.
\newblock In \emph{Proceedings of the 19th Conference of the European Chapter
  of the Association for Computational Linguistics (Volume 1: Long Papers)},
  pages 7080--7106. Association for Computational Linguistics.

\bibitem[{Wang et~al.(2023)Wang, Xie, Jiang, Mandlekar, Xiao, Zhu, Fan, and
  Anandkumar}]{wang2023voyager}
Guanzhi Wang, Yuqi Xie, Yunfan Jiang, Ajay Mandlekar, Chaowei Xiao, Yuke Zhu,
  Linxi Fan, and Anima Anandkumar. 2023.
\newblock \href {https://doi.org/10.48550/arXiv.2305.16291} {Voyager: An
  open-ended embodied agent with large language models}.
\newblock \emph{arXiv preprint arXiv:2305.16291}.

\bibitem[{Wang et~al.(2025)Wang, Mao, Fried, and
  Neubig}]{wang2025agentworkflow}
Zora~Zhiruo Wang, Jiayuan Mao, Daniel Fried, and Graham Neubig. 2025.
\newblock \href {https://proceedings.mlr.press/v267/wang25bx.html} {Agent
  workflow memory}.
\newblock In \emph{Proceedings of the 42nd International Conference on Machine
  Learning}, volume 267 of \emph{Proceedings of Machine Learning Research},
  pages 63897--63911. PMLR.

\bibitem[{Xu et~al.(2025)Xu, Liang, Mei, Gao, Tan, and Zhang}]{xu2025amem}
Wujiang Xu, Zujie Liang, Kai Mei, Hang Gao, Juntao Tan, and Yongfeng Zhang.
  2025.
\newblock \href
  {https://proceedings.neurips.cc/paper_files/paper/2025/hash/19909c36f51abc4856b4560aff3d36d6-Abstract-Conference.html}
  {{A-MEM}: Agentic memory for {LLM} agents}.
\newblock In \emph{Advances in Neural Information Processing Systems},
  volume~38.

\bibitem[{Yang et~al.(2026)Yang, Tian, Wang, Liu, Ye, and Chen}]{yang2026gse}
Chen Yang, Jiashuo Tian, Ziqi Wang, Xinyin Liu, Meiru Ye, and Junjie Chen.
  2026.
\newblock \href {https://arxiv.org/abs/2608.06153} {Learning globally reusable
  skills for coding agents}.
\newblock \emph{Preprint}, arXiv:2608.06153.

\bibitem[{Yang et~al.(2023)Yang, Song, Li, Zhao, Ge, Li, and
  Shan}]{yang2023gpt4tools}
Rui Yang, Lin Song, Yanwei Li, Sijie Zhao, Yixiao Ge, Xiu Li, and Ying Shan.
  2023.
\newblock \href
  {https://proceedings.neurips.cc/paper_files/paper/2023/hash/e393677793767624f2821cec8bdd02f1-Abstract-Conference.html}
  {{GPT4Tools}: Teaching large language model to use tools via
  self-instruction}.
\newblock In \emph{Advances in Neural Information Processing Systems},
  volume~36, pages 71995--72007. Curran Associates, Inc.

\bibitem[{Yao et~al.(2023)Yao, Zhao, Yu, Du, Shafran, Narasimhan, and
  Cao}]{yao2023react}
Shunyu Yao, Jeffrey Zhao, Dian Yu, Nan Du, Izhak Shafran, Karthik~R.
  Narasimhan, and Yuan Cao. 2023.
\newblock \href {https://openreview.net/forum?id=WE_vluYUL-X} {{ReAct}:
  Synergizing reasoning and acting in language models}.
\newblock In \emph{The Eleventh International Conference on Learning
  Representations}.

\bibitem[{Zhang et~al.(2026{\natexlab{a}})Zhang, Hu, Upasani, Ma, Hong,
  Kamanuru, Rainton, Wu, Ji, Li, Thakker, Zou, and Olukotun}]{zhang2026ace}
Qizheng Zhang, Changran Hu, Shubhangi Upasani, Boyuan Ma, Fenglu Hong,
  Vamsidhar Kamanuru, Jay Rainton, Chen Wu, Mengmeng Ji, Hanchen Li, Urmish
  Thakker, James Zou, and Kunle Olukotun. 2026{\natexlab{a}}.
\newblock \href {https://openreview.net/forum?id=eC4ygDs02R} {Agentic context
  engineering: Evolving contexts for self-improving language models}.
\newblock In \emph{The Fourteenth International Conference on Learning
  Representations}.

\bibitem[{Zhang et~al.(2024)Zhang, Tang, Wu, Wang, Shen, Hou, Tan, Li, Zhuang,
  and Lu}]{zhang2024agentpro}
Wenqi Zhang, Ke~Tang, Hai Wu, Mengna Wang, Yongliang Shen, Guiyang Hou, Zeqi
  Tan, Peng Li, Yueting Zhuang, and Weiming Lu. 2024.
\newblock \href {https://doi.org/10.18653/v1/2024.acl-long.292} {Agent-pro:
  Learning to evolve via policy-level reflection and optimization}.
\newblock In \emph{Proceedings of the 62nd Annual Meeting of the Association
  for Computational Linguistics (Volume 1: Long Papers)}, pages 5348--5375.
  Association for Computational Linguistics.

\bibitem[{Zhang et~al.(2026{\natexlab{b}})Zhang, Bu, Pan, Miao, Zhang, Wang,
  Ji, Tang, Li, and Tang}]{zhang2026ca3mem}
Zhenkui Zhang, Wendong Bu, Kaihang Pan, Bingchen Miao, Wenqiao Zhang, Guoming
  Wang, Wei Ji, Rui Tang, Juncheng Li, and Siliang Tang. 2026{\natexlab{b}}.
\newblock \href {https://doi.org/10.1609/aaai.v40i15.38300} {Evolving
  generalist virtual agents with generative and associative memory}.
\newblock \emph{Proceedings of the AAAI Conference on Artificial Intelligence},
  40(15):13006--13014.

\bibitem[{Zhao et~al.(2024)Zhao, Huang, Xu, Lin, Liu, and
  Huang}]{zhao2024expel}
Andrew Zhao, Daniel Huang, Quentin Xu, Matthieu Lin, Yong-Jin Liu, and Gao
  Huang. 2024.
\newblock \href {https://doi.org/10.1609/aaai.v38i17.29936} {{ExpeL}: {LLM}
  agents are experiential learners}.
\newblock \emph{Proceedings of the AAAI Conference on Artificial Intelligence},
  38(17):19632--19642.

\bibitem[{Zhong et~al.(2026)Zhong, Lu, Ning, Wan, Feng, Ao, Ribeiro, Dreyer,
  Ammirati, and Xiong}]{zhong2026skilllearnbench}
Shanshan Zhong, Yi~Lu, Jingjie Ning, Yibing Wan, Lihan Feng, Yuyi Ao, Leonardo
  F.~R. Ribeiro, Markus Dreyer, Sean Ammirati, and Chenyan Xiong. 2026.
\newblock \href {https://arxiv.org/abs/2604.20087} {{SkillLearnBench}:
  Benchmarking continual learning methods for agent skill generation on
  real-world tasks}.
\newblock \emph{Preprint}, arXiv:2604.20087.

\bibitem[{Zhou et~al.(2025)Zhou, Hua, Pan, Cheng, Wu, Yu, and
  Wang}]{zhou2025rulearena}
Ruiwen Zhou, Wenyue Hua, Liangming Pan, Sitao Cheng, Xiaobao Wu, En~Yu, and
  William~Yang Wang. 2025.
\newblock \href {https://doi.org/10.18653/v1/2025.acl-long.27} {{RuleArena}: A
  benchmark for rule-guided reasoning with {LLM}s in real-world scenarios}.
\newblock In \emph{Proceedings of the 63rd Annual Meeting of the Association
  for Computational Linguistics (Volume 1: Long Papers)}, pages 550--572.

\bibitem[{Zhou et~al.(2024)Zhou, Xu, Zhu, Zhou, Lo, Sridhar, Cheng, Ou, Bisk,
  Fried, Alon, and Neubig}]{zhou2024webarena}
Shuyan Zhou, Frank~F. Xu, Hao Zhu, Xuhui Zhou, Robert Lo, Abishek Sridhar,
  Xianyi Cheng, Tianyue Ou, Yonatan Bisk, Daniel Fried, Uri Alon, and Graham
  Neubig. 2024.
\newblock \href {https://openreview.net/forum?id=oKn9c6ytLx} {{WebArena}: A
  realistic web environment for building autonomous agents}.
\newblock In \emph{The Twelfth International Conference on Learning
  Representations}.

\end{thebibliography}

\appendix
\section{Prompt Templates and Intermediate Outputs}
\label{app:method-prompts}
This appendix reproduces the principal prompt templates used in the three
stages of \method and presents representative intermediate outputs from an
actual RuleArena NBA run.  Prompts are abbreviated only by removing repeated
formatting instructions; the displayed inputs, constraints, and output fields
match the implemented pipeline.
\subsection{Stage 1: Patch-Author Prompt and Example Patch}
\label{app:patch-induction-prompts}
Stage~1 contrasts a failed executor trajectory with a verifier-accepted teacher
trajectory and asks the patch author for one executable repair.  The frozen
executor then receives only the proposed \texttt{repair\_memory} and replays the
same source task.  A failed replay is returned to the author for refinement,
and the process stops after the first verifier pass or after $K=3$ attempts.
\paragraph{Patch-author prompt.}
The initial and refinement prompts share the same output contract.  On a
refinement attempt, the prompt additionally includes the previous repair and
the latest replay trajectory and asks the author to address the new first
decisive divergence rather than repeat the previous wording.

\begin{promptlisting}{lst:patch-author-prompt}{Stage 1---episode-patch author prompt (abbreviated).}
\small\raggedright
\noindent\textbf{Role:} You are the same weak solver writing a compact repair
for your own failed attempt.\par

\medskip
\noindent\textbf{Evidence:} You receive the original task, your failed
reasoning and answer, and a training-only answer-aware teacher trace that has
passed the task verifier.\par

\medskip
\noindent\textbf{Mission:} Locate the first decisive divergence and write one
executable episode-level repair for a fresh attempt. The repair may explain how
to interpret supplied rules, track state, order operations, perform
calculations, control search, verify intermediate results, or serialize the
answer.\par

\medskip
\noindent\textbf{Requirements:}\par
1. Write behavioral guidance, not an answer key.\par
2. Do not include the exact final answer, copy the teacher's solved
decomposition, or encode a case-specific lookup.\par
3. Make the repair compact and directly executable by the weak solver.\par
4. The replay solver will see only \texttt{repair\_memory}, not the teacher
trace or the remaining analysis fields.\par

\medskip
\noindent\textbf{Output:} Return valid JSON containing whether the failure is
repairable; summaries of the weak failure and teacher mechanism; a repair
strategy type; \texttt{repair\_memory}; concrete behavior constraints; an
adherence checklist; the recommended memory level; and confidence.\par
\end{promptlisting}

\begin{promptlisting}{lst:actual-instance-patch}{Actual Stage 1 patch and source-replay result. Text is reproduced verbatim apart from line wrapping and a shortened failure summary.}
\small\raggedright
\noindent\textbf{Case:} RuleArena NBA; two sequential player signings by the
same team (internal ID: \texttt{comp0\_003}).\par

\medskip
\noindent\textbf{Weak failure summary:} The solver concluded both operations
legal, failing to check the interaction between Room MLE usage and subsequent
exception eligibility.\par

\medskip
\noindent\textbf{Repair memory:} When evaluating multiple operations for a
team, track the team's salary cap status and exception usage sequentially.
After applying an exception like Room MLE, check if the team is now above the
cap and which exceptions remain available. Verify that any subsequent
operation can be accommodated either by remaining cap room (if below cap) or
by a currently available exception with sufficient amount. Do not assume a
second operation can use a different exception without checking the rule that
using Room MLE prohibits using Bi-annual, Non-Taxpayer MLE, or Taxpayer MLE in
the same salary cap year.\par

\medskip
\noindent\textbf{Source replay:} PASS (attempt 1 of 3).\par
\end{promptlisting}

\paragraph{Representative case.}
Listing~\ref{lst:actual-instance-patch} shows the first-attempt output for
a RuleArena NBA induction case, internally indexed as
\texttt{comp0\_003}, in which one team attempts two consecutive player
signings in the same salary-cap year.  The first signing requires the Room MLE
because the team's cap room is insufficient; evaluating the second therefore
requires carrying forward the updated team salary, remaining exception amount,
and exception eligibility.  The failed executor instead treated the two
operations independently.  The author converted this discrepancy into a
sequential state-tracking repair, which passed the native task verifier on its
first source replay.

\subsection{Stage 2: Shared-Mechanism Assessment and Example Group}
\label{app:compatibility-prompts}

Stage~2 uses semantic similarity only to retrieve a bounded candidate set.  It
then cross-replays the incoming source-validated patch and the retrieved
memories before asking a mechanism assessor whether they share one executable
correction mechanism.  A positive decision yields either a proposed expansion
scope containing an active skill and the incoming patch, or a coherent group of
residual patches.  No higher-level skill is authored in this stage.

\paragraph{Mechanism-assessment prompt.}
The controller supplies a replay-supported candidate group.  The model then
assesses whether the group admits one shared executable repair mechanism.
For a root proposal it may exclude mechanistically different outliers; for an
expansion it must decide whether the incoming patch follows the active skill's
exact mechanism.  Its output defines the proposed source group passed to
Stage~3, not the text of a higher-level skill.

\begin{promptlisting}{lst:skill-author-prompt}{Stage 2---shared-mechanism assessment prompt (abbreviated).}
\small\raggedright
\noindent\textbf{Role:} You determine whether several validated repairs follow
one shared executable mechanism. Do not author or rewrite a higher-level
skill.\par

\medskip
\noindent\textbf{Inputs:} Either several source-validated episode repairs
forming a replay-supported residual group, or one active parent skill plus a
new source-validated repair. The final episode repair is the new arrival.\par

\medskip
\noindent\textbf{Shared-mechanism assessment:}\par
1. For a root proposal, select the largest subset that shares one executable
correction mechanism and includes the new arrival. Exclude topical but
mechanistically different outliers instead of rejecting the whole group.\par
2. For an expansion, determine whether the incoming repair shares the parent's
exact execution mechanism. If not, abstain; do not broaden the parent into a
checklist of unrelated procedures.\par
3. If no qualifying coherent subset exists, return
\texttt{\{"coherent": false\}}.\par

\medskip
\noindent\textbf{Assessment requirements:}\par
1. Base the decision on the replay evidence and the exact execution procedure,
not topical or lexical similarity.\par
2. A selected root group must include the new arrival; remove mechanistically
different outliers rather than broadening the shared mechanism.\par
3. Describe the common mechanism only as evidence for grouping. Do not produce
\texttt{when}, \texttt{repair}, or \texttt{avoid\_when}.\par

\medskip
\noindent\textbf{Output:} When the group is coherent, return
\texttt{coherent}, the one-based sorted
\texttt{selected\_source\_indices}, and a concise
\texttt{shared\_mechanism\_summary}. Otherwise, return
\texttt{\{"coherent": false\}}.\par
\end{promptlisting}

\paragraph{Representative case.}
In the example below, three source-validated patches discuss different NBA
operations, but cross-replay supports a common procedural mechanism: process
operations in order and carry forward the updated team state.  The assessor
selects all three patches as one proposed source group.  Skill induction is
deferred to Stage~3.  The repair snippets and mechanism summary are shortened
for readability.

\begin{promptlisting}{lst:actual-grouping-case}{Representative Stage 2 mechanism assessment and residual group.}
\small\raggedright
\noindent\textbf{Episode repairs:}\par
1. Track salary-cap status and exception usage sequentially; update available
exceptions after using Room MLE.\par
2. Compute current salaries, check apron status, and apply the corresponding
traded-player limit.\par
3. Process a trade and signing sequentially; use the post-trade salary and
exception state for the signing.\par

\medskip
\noindent\textbf{Author output:} \texttt{coherent = true};
\texttt{selected\_source\_indices = [1, 2, 3]}.\par

\medskip
\noindent\textbf{Shared mechanism summary:} Process related operations in
order, update the affected team state after each operation, and evaluate the
next operation against that updated state rather than stale pre-operation
values.\par

\medskip
\noindent\textbf{Stage output:} Proposed residual source group
\texttt{[1, 2, 3]}; no higher-level skill has yet been authored.\par
\end{promptlisting}
\subsection{Stage 3: Skill Induction, Revision, and Consolidation}
\label{app:consolidation-prompts}

Stage~3 receives the proposed group produced by Stage~2 and first authors a
higher-level candidate skill over that scope.  For a residual group, it induces
a new skill; for an active-skill match, it authors an expanded version of the
parent skill.  The candidate is then replayed on every source in its proposed
scope.  The author makes at most $L=3$ attempts, including the initial
candidate, with later attempts guided by failed source replays.  Only a fully
supported scope is committed; partial support may form a separately revalidated
split, while insufficient support leaves the existing library unchanged.

\paragraph{Skill-authoring and replay-guided revision prompt.}
On the first attempt, the author receives the Stage~2 group, its shared-mechanism
summary, and the active parent skill when proposing an expansion, and renders
the mechanism as \texttt{when}, \texttt{repair}, and
\texttt{avoid\_when}.  A refinement attempt additionally includes the previous
candidate and anonymous per-source replay results.  The source set is locked so
that the author must repair an observed omission instead of hiding it by
dropping a failed source.

\begin{promptlisting}{lst:consolidation-revision-prompt}{Stage 3---skill-authoring and replay-guided revision prompt (abbreviated).}
\small\raggedright
\noindent\textbf{Role:} You author a higher-level candidate skill from a
mechanism-compatible source group and revise it when full-scope replay exposes
an omission.\par

\medskip
\noindent\textbf{Inputs:} The locked source repairs selected in Stage~2; their
shared-mechanism summary; and the active parent skill, if any. On a refinement
attempt, also receive the previous candidate, anonymous per-source pass/fail
results, failed solver outputs, and parsed predictions.\par

\medskip
\noindent\textbf{Mission:} On the initial attempt, express the shared mechanism
as one reusable and executable skill. On refinement, repair omissions exposed
by failed source replays while preserving the shared mechanism and claimed
source scope.\par

\medskip
\noindent\textbf{Requirements:}\par
1. The proposed source set is locked; preserve all source indices during
revision.\par
2. Do not solve any source problem, invent source facts, or hide a failed
source by narrowing the claim.\par
3. Put every execution step in a self-contained \texttt{repair}; use
\texttt{when} and \texttt{avoid\_when} only as applicability and exclusion
guards.\par
4. Do not copy source-specific answers, identifiers, or provenance into the
candidate.\par

\noindent\textbf{Output:} Return the locked source indices and the candidate
fields \texttt{when}, \texttt{repair}, and \texttt{avoid\_when}. The
controller, not the model, subsequently uses replay support to commit, split,
or reject the candidate.\par
\end{promptlisting}

\paragraph{Example induced candidate skill.}
Given the compatible residual group in
Listing~\ref{lst:actual-grouping-case}, the Stage~3 author converts its shared
mechanism into the following candidate representation.

\begin{promptlisting}{lst:actual-selected-candidate}{Representative Stage 3 candidate induced from the Stage 2 group.}
\small\raggedright
\noindent\textbf{When:} When evaluating multiple operations (trades, signings,
exceptions) for a team in the same salary-cap year.\par

\medskip
\noindent\textbf{Repair:} Process each operation sequentially. After each
operation, update the team's current salary, cap room, and available
exceptions. For the next operation, use the updated team salary and cap status
to determine eligibility and limits. Do not assume the previous operation's
state remains unchanged.\par

\medskip
\noindent\textbf{Avoid when:} When only a single operation is being evaluated,
or when operations are independent and do not affect each other's cap status.\par
\end{promptlisting}

\begin{promptlisting}{lst:actual-consolidation-case}{Actual Stage 3 replay, selection, and split-commit outcome.}
\small\raggedright
\noindent\textbf{Proposed scope:} Seven source instances.\par

\medskip
\noindent\textbf{Full-scope replay:} attempt 0, 5/7 passed; attempt 1, 6/7
passed (selected); attempt 2, 4/7 passed.\par

\medskip
\noindent\textbf{Selected version:} Six supported sources; one excluded source
(internal ID: \texttt{comp0\_034}).\par

\medskip
\noindent\textbf{Reduced-scope replay:} 6/6 passed.\par
\noindent\textbf{Transition:} \texttt{COMMIT\_SPLIT}.\par
\noindent\textbf{Preserved state:} The parent remains active and unchanged;
the excluded source retains its previous status under the parent.\par
\end{promptlisting}

\paragraph{Example committed split.}
Listing~\ref{lst:actual-consolidation-case} presents a separate expansion
example with seven proposed sources.  Across the three authored
versions, full-scope replay supported five, six, and four of the seven claimed
sources.  The controller selected the second version, narrowed its scope to the
six sources it supported, and replayed that reduced scope once more.  Successful
revalidation committed an independent sibling skill; the original parent and
the excluded source retained their previous states.
\FloatBarrier

\section{Experimental and Implementation Details}
\label{app:experimental-details}

This appendix records the implementation choices and controlled experimental
settings needed to reproduce the reported comparisons.

\subsection{Dataset and Evaluation Interfaces}
\label{app:evaluation-interfaces}

\begin{table}[t]
\centering
\small
\setlength{\tabcolsep}{4pt}
\renewcommand{\arraystretch}{1.08}
\begin{tabular}{lrrl}
\toprule
\textbf{Task} & \textbf{Induct.} & \textbf{Test} & \textbf{Metric} \\
\midrule
RuleArena Airline & 100 & 100 & strict accuracy \\
RuleArena NBA & 170 & 45 & strict accuracy \\
OpenExempt EC & 210 & 105 & macro-F1 \\
OpenExempt EV & 210 & 105 & macro-F1 \\
KOR-Bench Logic & 175 & 75 & strict accuracy \\
KOR-Bench Cipher & 175 & 75 & strict accuracy \\
\bottomrule
\end{tabular}
\caption{Construction and test cases.}
\label{tab:appendix-splits}
\end{table}

\paragraph{Split boundary.}
Each task type has an independent library.  RuleArena Airline uses Comp1 for
construction and Comp2 for evaluation; NBA uses Comp0 and Comp1 for
construction and Comp2 for evaluation.  OpenExempt and KOR-Bench use the
disjoint induction/test manifests summarized in
\autoref{tab:appendix-splits}.  Test tasks, labels, verifier feedback, and
test-time predictions are unavailable to patch induction, pattern authoring,
threshold selection, and library construction.  The library is serialized and
frozen before evaluation.

\paragraph{Parsing and verification.}
Every method is evaluated through the same task adapter within a model
configuration.  The adapter constructs the unchanged task request, parses the
model response into the benchmark's native prediction object, and invokes the
native verifier.  RuleArena and KOR-Bench record a binary correct/incorrect
outcome; any missing, malformed, or verifier-rejected final answer receives
zero credit.  OpenExempt retains the official structured prediction and
evaluator, so missing applicable citations/claims and unsupported
citations/claims both reduce macro-F1.  A provider failure is journaled and
retried, but a successfully returned response that fails parsing is an
incorrect task attempt and is not regenerated.

\paragraph{Construction-time replay.}
Source and cross-instance replays use the same adapter, solver model, task
prompt, parser, and verifier as the corresponding no-skill execution.  The only
request change is insertion of one candidate memory.  Replay records persist
the model and prompt versions, parsed prediction, verifier result, token usage,
finish reason, and a response reference.  Cached replay results are keyed by
the memory version and target episode, preventing a changed pattern from
inheriting evidence collected for an earlier version.
\FloatBarrier

\subsection{Models and Generation Settings}
\label{app:model-generation-settings}

\begin{table}[t]
\centering
\footnotesize
\setlength{\tabcolsep}{3.5pt}
\renewcommand{\arraystretch}{1.1}
\begin{tabular}{@{}>{\raggedright\arraybackslash}p{0.20\columnwidth}
                >{\raggedright\arraybackslash}p{0.46\columnwidth}
                >{\raggedright\arraybackslash}p{0.25\columnwidth}@{}}
\toprule
\textbf{Setting} & \textbf{Author $\rightarrow$ user}
& \textbf{Purpose} \\
\midrule
DS matched & DeepSeek-V4-Flash $\rightarrow$ same & matched model \\
Qwen matched & Qwen3.6-35B-A3B $\rightarrow$ same & matched model \\
Cross-scale & Qwen3.6-35B-A3B $\rightarrow$ Qwen3.5-9B & same family \\
Cross-family & GPT-5.4 $\rightarrow$ Qwen3.5-9B & family + scale \\
\bottomrule
\end{tabular}
\caption{Skill-author and skill-user configurations. Exact provider model IDs
are given in the surrounding text.}
\label{tab:appendix-models}
\end{table}

\paragraph{Frozen roles.}
The skill user is never updated by gradient descent or parameter editing.
Within a configuration, no-skill execution, all baseline methods, \method
source replay, cross replay, and final evaluation share the same user model.
The author model is used only for patch/pattern construction and does not
replace the task executor.  Teacher/reference generation is an
induction-only operation.
The provider identifiers are \texttt{deepseek-v4-flash},
\texttt{Qwen/Qwen3.6-35B-A3B}, \texttt{Qwen/Qwen3.5-9B}, and
\texttt{gpt-5.4-2026-03-05}.

\paragraph{Decoding.}
Patch and pattern authoring require JSON objects and use temperature $0$.
Pattern authoring has a maximum output budget of 8,192 tokens.  The
benchmark-specific Qwen/GPT pipelines use the limits in
\autoref{tab:appendix-token-budgets}; task execution and all replay calls for
a task share the same solver limit.  Qwen inference disables hidden thinking;
the task profile retains its declared sampling parameters, while all structured
author calls remain deterministic.  Provider calls use content-addressed
journals, a 600-second request timeout where supported, and at most three API
attempts by default.  OpenExempt construction uses up to six attempts because
of longer structured responses; provider failures never count as verifier
passes.

\begin{table}[t]
\centering
\small
\setlength{\tabcolsep}{4pt}
\renewcommand{\arraystretch}{1.08}
\begin{tabular}{lrr}
\toprule
\textbf{Benchmark} & \textbf{Patch author} & \textbf{Solver/replay} \\
\midrule
RuleArena & 8,192 & 24,576 \\
OpenExempt & 8,192 & 16,384 \\
KOR-Bench & 16,384 & 8,192 \\
\bottomrule
\end{tabular}
\caption{Maximum output tokens in the Qwen/GPT construction pipelines.}
\label{tab:appendix-token-budgets}
\end{table}

\paragraph{Main-table evaluation.}
\label{app:main-table-deployment}
Avg. is the unweighted mean of the six task scores.
SkillCAT and GSE evaluations in \autoref{tab:main-results} use frozen libraries.
DeepSeek NBA/EV deployment reevaluations use a 64K solver output limit; their averages combine these scores with the original 16K limits for Airline/EC and 8K for Logic/Cipher.
The Qwen Airline evaluations use self-hosted \texttt{Qwen3.6-35B-A3B-FP8} authors, with that model or \texttt{Qwen3.5-9B} as the user, and a 24K solver output limit.
Other methods retain their reported scores.

\paragraph{Semantic encoder.}
All \method retrieval views use
\texttt{text-embedding-3-small}.  Episode patches are embedded from the
accepted repair text.  Pattern versions are embedded from the ordered
\texttt{when}, \texttt{repair}, and \texttt{avoid\_when} fields.  Cosine
similarity is used only to rank a bounded candidate set; replay and the native
verifier determine compatibility.
\FloatBarrier

\subsection{Hyperparameters and Construction Budgets}
\label{app:hyperparameters-budgets}

\begin{table}[t]
\centering
\footnotesize
\setlength{\tabcolsep}{2.4pt}
\renewcommand{\arraystretch}{1.1}
\begin{tabular}{lcccccc}
\toprule
\textbf{Task family} & $K$ & $L$ & $R$ & $\gamma$ & $m$ & $W$ \\
\midrule
RuleArena & 3 & 3 & 4 & forward & 3 & 2 \\
OpenExempt & 3 & 3 & 4 & $2/3$ & 2 & 2 \\
KOR-Bench & 3 & 3 & 4 & $2/3$ & 2 & 2 \\
\bottomrule
\end{tabular}
\caption{Construction hyperparameters. $K$ is the maximum patch
generate--replay attempts, $L$ is the maximum pattern-authoring attempts
including the initial candidate,
$R$ is the semantic shortlist size, $\gamma$ is the active-skill
compatibility cutoff, $m$ is the minimum residual group/split size, and
$W$ is the maximum number of active-skill source witnesses used for
reverse replay. RuleArena requires the active skill to pass on the incoming
source before mechanism assessment; the other pipelines use the displayed
graded threshold.}
\label{tab:appendix-hyperparameters}
\end{table}

\paragraph{Bounded construction.}
Patch induction makes at most $K$ author calls and $K$ source replays per
initial executor failure, stopping at the first pass.  A new arrival retrieves
at most $R$ active skills; active-skill compatibility uses one forward replay
per candidate and at most two reverse witness replays for the selected
candidate.  Residual grouping retrieves at most $R$ patches and evaluates the
two directed replays for each retrieved pair.  A proposed abstraction makes at
most $L$ pattern-author calls and replays each version on every claimed source.
If a partial-support version is selected, one final replay is performed on the
reduced source set before a split can commit.

These are upper bounds rather than fixed call allocations: successful source
replay stops patch refinement, cached directed transfers are reused by their
content-addressed key, an author may abstain, and an arrival with no eligible
group remains residual without pattern authoring.  Every compared method
receives the same available construction cases shown in
\autoref{tab:appendix-splits}, but we preserve each method's native
construction loop.  Construction costs are reported in
\AppendixRef{app:construction-cost}.
\FloatBarrier

\subsection{Baseline Implementations and Fairness Controls}
\label{app:baseline-implementations}

\begin{table}[t]
\centering
\footnotesize
\setlength{\tabcolsep}{3pt}
\renewcommand{\arraystretch}{1.08}
\begin{tabular}{@{}>{\raggedright\arraybackslash}p{0.18\columnwidth}
                >{\raggedright\arraybackslash}p{0.21\columnwidth}
                >{\raggedright\arraybackslash}p{0.26\columnwidth}
                >{\raggedright\arraybackslash}p{0.24\columnwidth}@{}}
\toprule
\textbf{Method} & \textbf{Memory unit} & \textbf{Construction}
& \textbf{Test-time use} \\
\midrule
ExpeL
& Insight rules; successful trials
& Reflect over successes and failures
& Retrieve similar successful trials \\
\parbox[t]{0.18\columnwidth}{Reasoning\\Bank}
& Structured trajectory memories
& Self-judge and append each experience
& Retrieve nearest prior experience \\
ACE
& Structured global playbook
& Generator--Reflector--Curator loop
& Load selected frozen playbook \\
\parbox[t]{0.18\columnwidth}{Trace2\\Skill}
& Hierarchical skill document
& Local patches $\rightarrow$ hierarchical merge
& Preload frozen skill file \\
SkillCAT
& Core skill and capability nodes
& Contrastive edits; assessed updates; hierarchical merge
& Route to nodes and prerequisites \\
GSE
& Skills and relation graph
& Generalize edits; compare candidate and control replays
& Route with prerequisites and conflicts \\
\bottomrule
\end{tabular}
\caption{Standardized summary of the reproduced baseline pipelines.}
\label{tab:appendix-baselines}
\end{table}

\paragraph{ExpeL.}
We preserve its reflection retries over successful and failed trials, ordered
insight-rule updates, and task-conditioned retrieval of successful examples.
The benchmark's native verifier supplies the outcome signal in place of an
interactive environment reward.

\paragraph{ReasoningBank.}
After producing the current prediction, the method self-judges the completed
trajectory, extracts a structured success or failure memory, and appends it to
the bank.  The benchmark task text queries the nearest earlier experience, so
memory updates retain the method's causal online order.

\paragraph{ACE.}
We retain the Generator, Reflector, and Curator roles that iteratively evolve a
global structured playbook.  Benchmark context and native verifier feedback
replace environment-specific fields, and induction-only validation selects the
playbook frozen for evaluation.

\paragraph{Trace2Skill.}
The reproduction converts verifier-labeled local trajectories into patches and
hierarchically merges them into one reusable skill document.  Because the
evaluated executors do not dynamically open auxiliary files, the resulting
frozen skill file is preloaded at evaluation time.

\paragraph{SkillCAT.}
The reproduction extracts experiences from successful and failed source
trajectories, proposes edits to a seed skill, and retains updates that meet its
source-replay assessment threshold.  Hierarchical merging produces a shared
core and capability nodes.  At evaluation, a task-conditioned router selects
nodes and their prerequisites within a node budget; the solver receives the
core and selected node contents.

\paragraph{GSE.}
The reproduction maintains skills and their prerequisite/conflict relations.
It groups source edits, proposes a generalized bank update, and compares the
candidate bank against the current bank on the same source and affected
historical tasks.  Updates are accepted when aggregate replay performance
does not decrease.  Test-time routing selects applicable skills, includes
prerequisites, and resolves conflicts before composing the solver context.

\paragraph{Shared controls.}
Within each model configuration, all conditions use the same frozen skill
user, base task prompt, benchmark-provided task context, output parser, native
verifier, and test-case order.  Solver output limits and deployment-specific
settings are specified in \AppendixRef{app:model-generation-settings}.  Offline
libraries use only the designated construction split and are frozen before
evaluation.  No test label or test-time verifier feedback is rendered to a
skill author, retriever, curator, or executor.  Method-specific authoring,
retrieval, reflection depth, and memory structure are retained because these
are part of the compared algorithms.

\paragraph{Failure accounting and recovery.}
Every construction/evaluation request records its prompt/model version and
content hash.  Provider retries resume the same logical request; parse failures,
schema violations, and token-limit truncations are recorded as failures.  Resuming a run reuses only a
response whose content-addressed request matches, which prevents stale results
from crossing model, prompt, memory-version, or target-instance boundaries.
\FloatBarrier

\section{Additional Experimental Results}
\label{app:additional-results}

\subsection{Online Library Construction Outcomes}
\label{app:library-construction-outcomes}

\autoref{tab:nba-construction-outcomes} summarizes the final online library
constructed from the 61 source-validated RuleArena NBA episode patches used in
the matched DeepSeek experiment.  These construction statistics complement
test accuracy by exposing what the grouping and consolidation stages
actually retained.

\begin{table}[t]
\centering
\small
\setlength{\tabcolsep}{5pt}
\renewcommand{\arraystretch}{1.08}
\begin{tabular}{lr}
\toprule
\textbf{Construction outcome} & \textbf{Count} \\
\midrule
Source-validated episode patches & 61 \\
Active higher-level skills & 7 \\
Residual episode patches & 15 \\
Unique episode sources covered by active skills & 46 \\
Root-skill commits & 4 \\
Split-skill commits & 3 \\
Successful attachments to existing skills & 30 \\
Deployable memories after consolidation & 22 \\
\bottomrule
\end{tabular}
\caption{RuleArena NBA online construction outcomes.}
\label{tab:nba-construction-outcomes}
\end{table}

The final deployable library contains seven higher-level skills and fifteen
residual patches, or 22 memories for 61 admitted source repairs (36.1\% of the
unconsolidated count).  The active skills cover 46 unique source episodes, and
every committed version achieves complete replay support over its claimed
source set.  Consolidation therefore compresses the library while retaining
the source coverage of its validated repairs.

\subsection{Construction Cost and Caching}
\label{app:construction-cost}

\autoref{tab:nba-construction-cost} reports model-generation cost for the
matched DeepSeek RuleArena NBA setting.  We count each method's native
construction protocol and logical tokens before provider cache discounts,
including required source-trajectory generation but excluding
evaluation.  ReasoningBank instead constructs memory online; its entry counts
the judge and extraction calls made after predictions, but not the 45 task
solves.

\begin{table}[t]
\centering
\small
\setlength{\tabcolsep}{3pt}
\renewcommand{\arraystretch}{1.08}
\begin{tabular}{lrrl}
\toprule
\textbf{Method} & \textbf{Generations} & \textbf{Tokens} & \textbf{When paid} \\
\midrule
ExpeL & 559 & 5.00M & Offline \\
ReasoningBank & 90 & 0.82M & Online updates \\
ACE & 1,115 & 14.52M & Offline \\
Trace2Skill & 319 & 2.15M & Offline \\
\method & 952 & 7.72M & Offline \\
\bottomrule
\end{tabular}
\caption{Logged memory-construction cost on RuleArena NBA with
DeepSeek-V4-Flash. Tokens are logical tokens rather than monetary cost and are
rounded to two decimal places.}
\label{tab:nba-construction-cost}
\end{table}

The totals expose a meaningful cost--validation trade-off.  \method uses more
construction compute than ExpeL and Trace2Skill because candidate repairs and
their abstractions are repeatedly executed rather than accepted from text
alone, but it remains below ACE's iterative generator--reflector--curator
budget.  The additional replay calls validate both repair transfer and
the source coverage of each abstraction.

Replay results are content-addressed and reusable.  In the last two stages, 461
logical replay requests required 445 distinct generations.  Reusing seeded and
cached results reduced the recorded replay charge from 3.97M logical tokens to
2.76M billed tokens (30.6\%) without changing replay decisions.
\autoref{tab:nba-construction-cost} uses pre-cache logical tokens consistently
across methods.
\FloatBarrier

\end{document}